\documentclass{article} 
\usepackage{iclr2027_conference,times}

\usepackage{hyperref}
\usepackage{url}
\usepackage[margin=1in]{geometry}
\usepackage{amsmath,amssymb,amsthm}
\usepackage{mathtools}
\usepackage{bm}
\usepackage{algorithm}
\usepackage{algorithmic}
\usepackage{booktabs}
\usepackage{tikz}
\usepackage{hyperref}
\usepackage{cleveref}
\crefname{appendix}{appendix}{appendices}
\Crefname{appendix}{Appendix}{Appendices}
\usepackage{natbib}
\usepackage{enumitem}
\usepackage{xcolor}

\usepackage{graphicx}
\usepackage{subcaption}
\usepackage{booktabs}

\usepackage{xcolor}
\usepackage{colortbl}
\definecolor{professionflip}{RGB}{232,242,252}

\title{Constructive Causal Abstraction \\ and A Case Study in Mechanistic Interpretability}

\author{Yuchen Zhu \\
Department of Computer Science\\
University College London \\
\texttt{yuchenz@spotify.com} \\
\And
Sergio Hernan Garrido Mejia \\
Max Planck Institute for Intelligent Systems \\
\texttt{sgarridomejia@gmail.com} \\
\AND
Dominik Janzing \\
Amazon \\
\texttt{janzind@amazon.de}
}

\usepackage{amsmath,amsfonts,bm}

\def\eqref#1{equation~\ref{#1}}

\crefname{assumption}{assumption}{assumptions}
\Crefname{assumption}{Assumption}{Assumptions}

\def\1{\bm{1}}

\DeclareMathAlphabet{\mathsfit}{\encodingdefault}{\sfdefault}{m}{sl}
\SetMathAlphabet{\mathsfit}{bold}{\encodingdefault}{\sfdefault}{bx}{n}

\newtheorem{theorem}{Theorem}
\newtheorem{proposition}[theorem]{Proposition}

\newtheorem{definition}{Definition}
\newtheorem{assumption}{Assumption}
\newtheorem{remark}{Remark}

\newcommand{\R}{\mathbb{R}}
\newcommand{\N}{\mathcal{N}}
\newcommand{\X}{\mathcal{X}}
\newcommand{\bx}{\bm{x}}
\newcommand{\bn}{\bm{n}}
\newcommand{\bA}{\bm{A}}
\newcommand{\bP}{\bm{P}}
\newcommand{\bq}{\bm{q}}
\newcommand{\bp}{\bm{p}}
\newcommand{\bi}{\bm{i}}
\newcommand{\bs}{\bm{s}}

\newcommand{\bff}{\bm{f}}
\newcommand{\bI}{\bm{I}}
\newcommand{\barA}{\bar{\bm{A}}}
\newcommand{\barx}{\bar{\bm{x}}}
\newcommand{\barn}{\bar{\bm{n}}}
\newcommand{\bari}{\bar{\bm{i}}}
\newcommand{\bars}{\bar{\bm{s}}}
\newcommand{\barf}{\bar{\bm{f}}}
\newcommand{\Ker}{\operatorname{Ker}}

\newcommand{\pa}{\mathrm{pa}}
\newcommand{\id}{\mathrm{id}}

\usepackage{cleveref}

\iclrfinalcopy 
\begin{document}

\maketitle
\begin{abstract}
Existing frameworks in causal abstraction first propose a high-level causal model, and then verify whether its interventional effects agree with those of a given micro-level model - the two models are declared causally consistent when the discrepancy is zero.
However, proposing a consistent macro causal model is difficult, because macro causal relations are almost never well-defined, since distinct micro-level interventions associated with the same apparent macro intervention can produce different effects. Our central insight is that causally consistent macro-level models can be obtained \textit{constructively}: ill-defined macro interventions can be interpreted as missing variables on the macro level, and can be accounted for by introducing a complementary variable $q$ that retains omitted causally relevant state. In the linear case, \(q\) has a unique form and we provide sufficient conditions for valid implementations to exist; in the nonlinear case, we provide an ODE-based algorithm for constructing implementations.
We call this framework \textit{Constructive Causal Abstraction} (CCA), and use CCA to formulate the difference between machine-learning model auditing and editing. With the Bias-in-Bios biographies dataset, we demonstrate that CCA allows practitioners to articulate assumptions when detecting the model's gender bias. Finally, CCA resolves a well-known paradox known as `interpretability illusion' which points to a mismatch between the subspace for editing and the one where the model implements a signal-to-output pathway.

\end{abstract}

\section{Introduction}
\label{sec:intro}

Scientific reasoning routinely operates at multiple levels of resolution.  
A macroeconomist models sector-level aggregates, not individual firms; a neuroscientist reasons about brain regions, not single synapses; an AI safety researcher targets high-level capabilities of a neural network, not individual weights.
In each case, practitioners intervene on a coarse-grained \emph{macro} model and expect the outcome to be predictable from that model alone, without tracking every microscopic degree of freedom.

{Causal abstraction} \citep{rubenstein2017causalconsistencystructuralequation,beckers2019abstracting,rischel2020category,kekic2024targeted} formalises this expectation.
A high-level model $\bar{M}$ is an exact transformation of a low-level model $M$ if there exists a mapping $\tau$ between their variable spaces such that a commutative diagram holds: intervening at the micro level and then aggregating yields the same outcome as aggregating first and then intervening at the macro level.

However, existing frameworks are \textit{verificational}: they assume the macro model is given a priori, which implicitly entails that effect of interventions are well-defined, and they check for consistency with the micro model. If the proposed macro model is not consistent, the existing frameworks do not provide guidance on the constructive question: 

\begin{quote}
\emph{Given a micro-level causal model and a set of macro-variables aggregated from the micro-variables, how do we come up with a macro model that is an exact transformation of the micro model?}
\end{quote}

The hardness of the problem arises because the aggregation map $\bp:\X\to\bar\X$ is many-to-one: for a given macro-intervention $\bari$, there are generically infinitely many micro-level realisations.
Unless further structure is imposed on the macro model, different realisations produce different macro-level causal effects, making the macro-intervention \emph{ill-defined}. In other words, there is an \textit{implementation gap}.

\paragraph{Macro level noise.}
In a structural causal model (SCM), exogenous noise variables encode variation not explained by the causal mechanisms.
When micro variables are aggregated into macro variables, the micro exogenous noise and the ambiguity of implementations must be coherently aggregated; otherwise the macro model has no well-defined noise, and no well-defined counterfactuals.
We show that the missing ingredient is a \emph{complementary variable} $\bq$: a function of the micro state and noise that encodes information discarded by $\bp$.
Specifying $\bq$ pins down which micro-realisations are admissible, collapsing the set of valid implementations to those that all produce the same macro-level causal effect.

\paragraph{Constructive Causal Abstraction.}Specifying the macro level noise $\bq$ along with the aggregation map $\bp$ can be viewed as projecting micro-level variables onto a subset of coordinates under a coordinate transformation of the underlying micro variable system; exact transformation arises by construction, since all we have done is a coordinate transformation. Therefore, we term our framework \emph{constructive causal abstraction} (CCA).

Our contributions are three-fold.
\begin{enumerate}
\item \textbf{Constructive foundation.}
We achieve causal exact transformation by construction, through formulating causal abstraction as a change of coordinates followed by a projection from the micro model onto macro variables $\bp$ and macro noise variable $\bq$.
This perspective provides the necessary foundation for reliably controlling micro-level systems through macro-level interventions (\Cref{sec:formulation}).

\item \textbf{Well-defined macro interventions and existence of implementations.}  
We point out and resolve ill-defined macro models due to ambiguous interventions, formally prove well-definedness of causal effects under our framework (\Cref{thm:linear-welldef}, \Cref{thm:nonlinear-welldef}). For linear causal models with linear aggregation maps, $\bq$ admits a unique-form expression and we establish mild sufficient conditions for existence of micro-level implementations of any macro shift intervention (\Cref{sec:linear}, \Cref{thm:linear-existence}). For nonlinear causal models, we derive an ODE-based algorithm that constructs valid implementations (\Cref{sec:nonlinear}).
\item \textbf{Case study in mechanistic interpretability.}
Applying our theoretical framework to model auditing and editing, we showcase that it helps capture distributional assumptions in the well-studied sensitive-attribute auditing scenario (\Cref{sec:bias-in-bios}) and demonstrate that it resolves a well-known paradox in mechanistic interpretability through an example in Indirect Object Identification (\Cref{sec:ioi-experiment}). 
\end{enumerate}

\section{Related Work}
\label{sec:related}

\paragraph{Causal abstraction.}
\citet{rubenstein2017causalconsistencystructuralequation} is first to define exact
transformations between SCMs.
\citet{beckers2019abstracting} strengthen this account, while
\citet{rischel2021compositional} develop a category-theoretic treatment of
abstraction error.
These accounts are verificational: given a micro model, macro model,
and cross-level maps, they assess whether the abstraction is
consistent. We take a constructive perspective: starting from a micro
model and macro variables, we derive the complementary structure needed
to make macro-interventions well-defined.
\citet{kekic2024targeted, zhu2024unsupervised} also construct abstractions, jointly
learning macro variables, intervention maps, and mechanisms from
interventional data; their identifiability results require conditions
including linearity, independent high-level noise, and sufficient
intervention coverage.
Our work addresses a logically prior question: when the macro variables
are specified, what makes an intervention on them well-defined?
We expose the ambiguity among micro-level implementations and resolve it
through the macro noise $\bq$, without additional assumptions, providing
a general foundation upon which further learning or identifiability
constraints may be imposed.
Finally, \citet{zhu2024meaningful} identify ambiguity in macro-level
interventions and illustrate resulting structural paradoxes, while
our work constructively resolves that ambiguity.

\paragraph{Causal coarsening and emergence.}
\citet{chalupka2016multi} study causal feature learning from a multi-resolution perspective although they work solely in the discrete case, and \citet{hoel2013quantifying} quantify the ``causal emergence'' that arises when macro models have more predictive power than micro models.
\citet{shalizi-what-is-a-macro-state} analyses when macroscopic variables can be treated as autonomous dynamical systems.
These works ask \emph{when} macro variables have autonomous causal dynamics; we address the complementary question of \emph{how} to construct them.

\paragraph{Mechanistic Interpretability.} 
Neural networks are studied through interventions on macro variables.
Sensitive-attribute audits ask whether demographic information is encoded
or causally used
\citep{dearteaga2019bias,vig2020causal,ravfogel2020null};
interchange interventions test causal abstractions
\citep{geiger2021-causal-abstraction4nn}; activation steering modifies
behaviour through learned directions
\citep{turner2023activation,li2023inference,zou2023representation};
and locate-then-edit methods modify factual associations
\citep{meng2022locating,meng2023mass}. Their conclusions can depend on the intervention's implementation.
\citet{hase2023does} find successful fact edits need not occur where
causal tracing localises the fact, while \citet{makelov2024subspace}
show that patches can act through dormant pathways.
\citet{wu2024reply} dispute interpreting these cases as ``illusions'',
citing training and evaluation choices.
Similar sensitivities affect activation patching and circuit ablation
\citep{zhang2024towards,miller2024transformer}, while
\citet{tan2024analysing} find that steering effects vary across inputs
and distributions. Meanwhile, CCA formalises one source of sensitivity as \emph{implementation ambiguity}:
different micro-level realisations of the same purported macro-intervention
need not agree.
Requiring implementations to preserve the complementary variable $\bq$
makes the macro-level effect well-defined and implementation-invariant.
Articulating $\bq$ lets practitioners specify preservation constraints
defining a well-defined macro model.
\section{Preliminaries}
\label{sec:prelim}

We build on \citet{peters2017elements} to define the class of \textbf{structural causal models} we consider in this paper.
\begin{definition}[SCM without measure]
\label{def:scm}
A structural causal model (SCM) is a tuple $(\mathcal{K}, \mathcal{J}, \X, \N, \bff)$ where $\mathcal{K}$ is a finite index set of endogenous variables; $\mathcal{J}$ is a finite index set of exogenous variables; $\X = \prod_{k\in\mathcal{K}} \X_k$ and $\N = \prod_{j\in\mathcal{J}} \N_j$ are measurable product spaces; and $\bff:\X\times\N\to\X$ is a measurable function called the causal mechanism.
\end{definition}

We do not assume a probability measure over $\N$ and we do not assume mutual independence of the exogenous variables (the model is semi-Markovian in general).
This allows our results to apply to a wide class of causal models.

We write $\bs:\N\to\X$ for the solution function satisfying $\bs(\bn) = \bff(\bs(\bn), \bn)$ for all $\bn\in\N$.

\begin{assumption}[Unique solutions]
\label{assp:unique-solutions}
For every $\bn\in\N$, the equation $\bx = \bff(\bx,\bn)$ has a unique solution.
The same holds for all intervened structural equations considered in this paper.
\end{assumption}
This assumption is a pointwise, measure-free counterpart of unique
solvability \citep[Definition~3.3 and Theorem~3.6]{bongers2021foundations}.
Since unique solvability need not be preserved under intervention
\citep[Section~3.4]{bongers2021foundations}, we impose it
for every intervention considered here. 
Note that we only require that a solution exists, which is always true for directed acyclic graphs (DAGs), but we do not assume that the causal models considered must form DAGs.

\begin{definition}[Intervention]
\label{def:intervention}
An intervention on variables $X_{[k]}$ is a map $i_{[k]}:\X_{[k]}\to\X_{[k]}$.
It transforms the structural equation from $f_{[k]}(X_{\pa([k])}, N_{[k]})$ to $i_{[k]} \circ f_{[k]}(X_{\pa([k])}, N_{[k]})$.
The identity map $\id$ represents no intervention.
\end{definition}

This covers shift interventions ($i_k(x) = x + c$) and do-interventions ($i_k(x) = c$) as special cases. To the best of our knowledge, we are the first to directly formulate interventions as maps on variable spaces that act on
structural mechanisms by post-composition. This
is a special case of the soft interventions defined by
\citet[Definition~10]{geiger2025causal} and
\citet[Definition~6]{massidda2023causal}, with the replacement mechanism being
$\tilde f_{[k]} = i_{[k]} \circ f_{[k]}$.

The \textbf{aggregation map} is a function $\bp: \X \to \bar\X$, where $\bar\X$ the space of macro variables. 
We write $\upsilon:\mathcal{K}\to\bar{\mathcal{K}}$ for the index assignment; $\upsilon(k) = \bar{k}$ indicates that micro-variable $k$ contributes to macro-variable $\bar{k}$.
\begin{assumption}[Block-decomposable aggregation]
\label{assp:aggregation}
The aggregation map $\bp:\X\to\bar\X$ with $\bar\X = \R^{\bar k}$ decomposes as $\bp = (\bp_1,\ldots,\bp_{\bar k})$, and each $\bp_{\bar k}:\X_{\upsilon^{-1}(\bar k)}\to\bar\X_{\bar k}$ depends only on the micro-variables assigned to macro-variable $\bar k$.
\end{assumption}
Crucially, what distinguishes our work from existing works is that we do not assume a priori that interventions on macro variables, and by extension, macro models, are well-defined. 

\section{Well-defined Macro Causal Model}
\label{sec:formulation}
\paragraph{Motivating Example.} Consider a retail company with $K$ products, each with sales volume $\mathbf{X}$ and unit price $\alpha_k$.
Total revenue is $Y = \sum_k \alpha_k X_k + N_Y$.
The stakeholder tracks a macro-variable $\bar X = \sum_k X_k$ (total category sales) and wishes to intervene: ``Increase category sales by $c$.'' This macro-intervention is ill-defined.
If we allocate the sales increase $c$ uniformly, and so each product's sales increases by $\delta_k = \frac{c}{K}$, then $Y$ shifts by $\frac{c}{K}\cdot(\sum_k \alpha_k)$. Meanwhile, concentrating the change on product 1 shifts $Y$ by $c\cdot\alpha_1$.
Unless all prices are equal, the two implementations produce different macro-level causal effects on revenue, even though both realize the same shift in $\bar X$. The ambiguity arises because the aggregation map $\bp: \mathbf{X} \mapsto \sum_k X_k$ discards information about the \emph{composition} of sales---information that is causally relevant to profit when products are sold at different prices. Therefore, in this case we allow ourselves to construct aggregation mappings to represent macro-level noise variables, given by the map $\bq: \mathcal X \times \mathcal N \to \bar{\mathcal N}$, with $\bar {\mathcal N} = \mathbb R^{\bar k}$. $\bq$ and $\bar X$ then together fully determines $Y$, thus resolving ambiguity; \emph{a priori}, an intervene on $\mathbf{X}$ may reflect on both $\bar X$ and $\bq$. The class of well-defined interventions may emerge by constraining to those which keep $\bq$ constant.

An implementation must only change the intervened macro variable and its descendants, but not the macro noise $\bq$.
\begin{definition}[Implementation]
\label{def:implementation}
Given a micro-model $M$, aggregation maps $\bp$ and $\bq$, and a macro-intervention $\bari_{\bar k}$ on $\bar X_{\bar k}$, an \emph{implementation} of $\bari_{\bar k}$ with respect to $(\bp, \bq)$ is a micro-level intervention $i_{\upsilon^{-1}(\bar k)}:\X_{\upsilon^{-1}(\bar k)}\to\X_{\upsilon^{-1}(\bar k)}$ s.t.:
\begin{enumerate}[leftmargin=*]
\item \textbf{Macro-target:} $\bp_{\bar k} \circ i_{\upsilon^{-1}(\bar k)} = \bari_{\bar k} \circ \bp_{\bar k}$.
\item \textbf{Noise preservation:} $\bq(\bs^{(\bi)}(\bn), \bn) = \bq(\bs(\bn), \bn)$ for all $\bn\in\N$.
\end{enumerate}
\end{definition}

Criterion~1 ensures the implementation achieves the desired macro-shift.
Criterion~2 ensures the complementary information is preserved, preventing off-target effects at the macro level.

\begin{definition}[Well-defined macro-intervention]
\label{def:welldef}
A macro-intervention $\bari_{\bar k}$ is \emph{well-defined} if there exists at least one implementation, and for all implementations $\bi, \bi'$ of $\bari_{\bar k}$, $\bp \circ \bs^{(\bi)}(\bn) = \bp \circ \bs^{(\bi')}(\bn)$ for all $\bn\in\N$.
\end{definition}

\begin{definition}[Well-defined macro causal model]
\label{def:welldef-model}
Given a micro causal model and an aggregation map $\bp$, a {macro causal model}  exists if there is a function $\bq:\X\times\N\to\bar\N$ and a macro structural function $\barf:\bar\X\times\bar\N\to\bar\X$ such that:
\begin{enumerate}[leftmargin=*]
\item Every macro-state can be reached from every other via some well-defined macro-intervention.
\item For every implementation $\bi$ of any macro-intervention $\bari$: $\bp\circ\bs^{(\bi)}(\bn) = \bars^{(\bari)}(\bq(\bs(\bn),\bn))$ for all $\bn$.
\end{enumerate}
\end{definition}


\noindent
\parbox[c]{0.57\linewidth}{%
Criterion 2 parallels the exact transformation condition of
\citet{rubenstein2017causalconsistencystructuralequation},
but requires consistency at the counterfactual rather than
interventional level, as illustrated by the diagram to the right.
}\hfill
\parbox[c]{0.40\linewidth}{%
\centering
\begin{tikzpicture}[auto, font=\small]
  \node (N) at (0,0.7) {$\N$};
  \node (X) at (1.7,0.7) {$\X$};
  \node (Nbar) at (0,1.7) {$\bar\N$};
  \node (Xbar) at (1.7,1.7) {$\bar\X$};

  \draw[->] (N) to node {$\bs^{(\bi)}$} (X);
  \draw[->] (N) to node[left, font=\scriptsize]
    {$\bq\circ(\bs\times\id)$} (Nbar);
  \draw[->] (X) to node[right] {$\bp$} (Xbar);
  \draw[->] (Nbar) to node {$\bars^{(\bari)}$} (Xbar);
\end{tikzpicture}%
}
\par

\paragraph{Interpretation.} $\bq$ must encode all ambiguities of the implementations unseen by the aggregation map. Provided $\bq$ is preserved, all implementations have consistent macro-level effects. If $\bq$ is too simple, then ill-defined intervention effects return; if $\bq$ is too rich (i.e. $\bq = (\bx, \bn)$), then no valid implementations achieve a non-zero macro change exists. In practice, we require $\bar{\mathcal N}$ to have the same dimensionality as $\bar{\mathcal X}$.

\section{Linear Case: Unique Macro Noise Form}
\label{sec:linear}

\paragraph{Set up.} Consider a linear micro SCM: $\bx = \bA\bx + \bn$, for which we assume solvable (\Cref{assp:unique-solutions}) with solution $\bs(\bn) = (I - \bA)^{-1}\bn$.
The aggregation map is a full-rank linear map $\bP\in\R^{\bar k \times k}$.
We seek a linear complementary variable $\bq = \begin{pmatrix} \bq_x & \bq_n \end{pmatrix}$ mapping $(\bx, \bn)$ to $\barn$, such that $\bar{\bx} = \bP \bx$ and $\bar{\bn} = \bq (\bx, \bn)$ together satisfies some linear causal model with structural equation $\bar A$, that is 
\begin{align}
    \bar{\bx} &= \bar{A} \bar{\bx} + \bar{\bn} \label{eq:macro-form}
\end{align}

\paragraph{Deriving the Complementary Variable.}
Noting that $\bar{\bx}$ can be obtained from the micro-level cause variables by $\bar{\bx} = \bP \bx = \bP (A \bx + \bn) 
=\bP \begin{pmatrix}  A & I_K\end{pmatrix} \begin{pmatrix}\bx \\ \bn\end{pmatrix}$. We would get the desired form \cref{eq:macro-form} if we can factorise the r.h.s. through $\begin{pmatrix} \bP & 0 \\ \bq_x & \bq_n \end{pmatrix}$. That is, consider the mapping $\bP \begin{pmatrix} A & I_K\end{pmatrix} 
\begin{pmatrix}
\bP & \mathbf{0}\\
\bq_x & \bq_n\end{pmatrix}^{-1}
\begin{pmatrix}
\bP & \mathbf{0}\\
\bq_x & \bq_n\end{pmatrix}
\begin{pmatrix}\bx \\ \bn \end{pmatrix}$.
\textit{A priori}, $\begin{pmatrix}
\bP & \mathbf{0}\\
\bq_x & \bq_n\end{pmatrix}^{-1}$ is not a function; instead it is a mapping from the image to the preimage of $\begin{pmatrix}
\bP & \mathbf{0}\\
\bq_x & \bq_n\end{pmatrix}$. A necessary condition for attaining $\cref{eq:macro-form}$ is that $\bP \begin{pmatrix} \bA & I \end{pmatrix} \begin{pmatrix} \bP & 0 \\ \bq_x & \bq_n \end{pmatrix}^{-1}$ is a well-defined function, which requires:
\begin{equation}
\label{eq:kernel-condition}
\Ker\begin{pmatrix} \bP & 0 \\ \bq_x & \bq_n \end{pmatrix} \subseteq \Ker\left[\bP\begin{pmatrix} \bA & I \end{pmatrix}\right].
\end{equation}

\begin{proposition}[Unique form for $\bq$]
\label{prop:q-form}
\Cref{eq:kernel-condition} and the macro model form \Cref{eq:macro-form} together imply:
\begin{equation}
\label{eq:q-linear}
\boxed{\bq = \begin{pmatrix} \bP\bA - \barA\bP & \bP \end{pmatrix}}
\end{equation}
for a macro-level causal matrix $\barA\in\R^{\bar k \times \bar k}$ chosen by the practitioner.
\end{proposition}

Proof is in \ref{app:proof-linear-q}.

\begin{remark}[Role of $\barA$]
\label{rem:barA}
The choice of $\barA$ determines both the macro-level causal mechanism and,
through \Cref{eq:q-linear}, the complementary variable $\bq$. How $\barA$ should be chosen depends on the practitioner's objective.
For control, the practitioner may specify a desired macro-level response, and
the resulting $\bq$ characterises the micro-level implementations required to
realise it, when such implementations exist.
For auditing, the purpose is to understand the behaviour of the macro model under the distribution of feasible micro-level interventions determined by the set-up of the practitioner. $\barA$ should describe the effect of changing the macro cause variable on the outcome averaged uniformly over choices of micro-level implementations - this choice of $\barA$ is given by setting $\bq$ to be uncorrelated with values of $\barA$. As pointed out by \citet{zhu2024meaningful}, this choice of $\bq$ corresponds to a stochastic intervention which leaves the ambiguity of implementations in their natural distribution, that is, we implement a macro intervention $\text{do} (\bar X := \bar x)$ by replacing the generating mechanism of $\mathbf X$ with $\text{do}(\mathbf X \sim \rho_{\mathbf X| \bp(\mathbf{X}) = \bar x})$, where $\rho$ is the observational measure of $\mathbf X$.
\end{remark}

\paragraph{Well-defined interventions.} We now establish that $\bq$ renders macro-causal effects well-defined.
\begin{theorem}[All macro-interventions are well-defined]
\label{thm:linear-welldef}
Under \Cref{assp:unique-solutions,assp:aggregation}, given the linear SCM $\bx = \bA\bx + \bn$ with a full-rank linear aggregation map $\bP$ and $\bq$ as in~\eqref{eq:q-linear}, every macro-intervention $\bari$ that has at least one implementation is well-defined.
For any implementation $\bi$ of $\bari$:
\[
\bp\circ\bs^{(\bi)}(\bn) = \bars^{(\bari)}(\bq(\bs(\bn), \bn)), \;\;\;\;\; \text{with } \bars^{(\bari)} = (I - \barA)^{-1}
\]
\end{theorem}
Proof is in \Cref{app:proof-linear-welldef}.

\paragraph{Motivating example revisited.} If we assert a macro-level causal model from total sales volume $\bar X = \sum_k X_k$ to total revenue $Y$, we can express the micro-level structural equation $Y = \sum_k \alpha_k X_k + N_Y$ in the new coordinate system of $\bar X$ and $\bq$: $    Y = \beta \underbrace{\sum_k X_k}_{\bar X} + \underbrace{\sum_k (\alpha_k - \beta) X_k + N_Y}_{\bq(\mathbf X, N_Y)}$. Note that we have to commit to a macro coefficient $\beta$. Now implementing a macro intervention $\text{do}(\bar X := \bar X + \bar \delta)$ requires a change on $\mathbf X$ to attain the $\delta$ change on $\bar X$ \emph{as well as} keeping $\bq(\mathbf X, N_Y)$ fixed; for instance, a micro-level shift intervention $\text{do}(\mathbf X := \mathbf X + \boldsymbol{\delta})$ which implements $\text{do}(\bar X := \bar X + \bar \delta)$ in the given macro-level model must satisfy 1. $\sum_k \delta_k = \bar \delta$, and 2. $\sum_k (\alpha_k - \beta) \delta_k$. With any valid $\boldsymbol{\delta}$ as implementation, we will clearly always increase $Y$ by $\beta \bar\delta$ when $\bar X$ increases by $\bar\delta$.

\paragraph{Existence of Implementations.}
To establish when valid implementations exist, we need two additional mild conditions, the first requires we are compressing significantly from high to low dimensions, and the second is a technical condition which is typically satisfied because $P$ and $A$ usually arise independently. 
\begin{assumption}[High-dimensional micro-state]
\label{assp:dim}
The micro-state dimension satisfies $k \geq 2 \bar k$.
\end{assumption}

\begin{assumption}[Linear independence]
\label{assp:lindep}
The rows of $\bP$ are linearly independent of those of $(\bP\bA - \barA\bP)(I - \bA)^{-1}$.
\end{assumption}

\begin{theorem}[Existence of macro shift interventions]
\label{thm:linear-existence}
Under \Cref{assp:unique-solutions,assp:aggregation,assp:dim,assp:lindep}, for the linear SCM with $\bq$ as in~\eqref{eq:q-linear}, every macro shift intervention $\bari(\barx) = \barx + \bar{\bm c}$ has at least one implementation.
\end{theorem}
Proof is in \Cref{app:proof-linear-existence}.

\begin{remark}
Since shift interventions can map any macro-state to any other, \Cref{thm:linear-welldef,thm:linear-existence} together establish that $\bq$ gives rise to a well-defined macro causal model in the sense of \Cref{def:welldef-model}.
\end{remark}

\section{Nonlinear Case: Numerical scheme for implementation estimation}
\label{sec:nonlinear}

For a general nonlinear SCM $\bx = \bff(\bx, \bn)$, 

we seek $\bq$ such that $\bp\circ\bff\circ(\bp^*, \bq)^{-1}$ is a well-defined function $\barf$, where $\bp^*(\bx, \bn) = \bp(\bx)$.
The complementary variable must satisfy:
\begin{equation}
\label{eq:q-nonlinear}
\bp\circ\bff(\bx, \bn) = \barf(\bp(\bx), \bq(\bx, \bn)) \quad \text{for all } (\bx, \bn).
\end{equation}

\paragraph{Well-defined interventions.} While a unique form for $\bq$ is generally unavailable, the well-definedness result extends:

\begin{theorem}[Well-defined macro-interventions, nonlinear case]
\label{thm:nonlinear-welldef}
Suppose $\bq$ satisfies~\eqref{eq:q-nonlinear} and \Cref{assp:unique-solutions} holds.
Then for every macro-intervention $\bari$ and all implementations $\bi, \bi'$ of $\bari$ with respect to $(\bp, \bq)$ have the same effect, that is:
$\bp\circ\bs^{(\bi)}(\bn) = \bp\circ\bs^{(\bi')}(\bn)$
and we may define $\bars^{(\bari)}(\bq(\bs(\bn), \bn)):=\bp\circ\bs^{(\bi)}(\bn) = \bp\circ\bs^{(\bi')}(\bn)$.
\end{theorem}
Proof is in \Cref{app:proof-nonlinear-welldef}.

\paragraph{Implementations as ODE solution.}
We realise that the micro-level solution function under an implementation can be expressed as a solution to an ODE.
Denote a macro-shift-intervention direction by a unit vector $\bari$ and let $T \bari$ be the total shift intervention. Furthermore, let $\bI(\bx, T)$ denote the implementation of $T \bari$ in the micro-level; $\bI(\bx, t)$ parameterises a family of implementations which implement $t \bari$, $t \in [0, T]$.

For a given micro shift intervention $\bI(\bx, t)$, let $\bs(\bn, t)$ be the solution to the intervened micro causal model, then
\begin{align}
    \bs(\bn, t) &= \bff(\bs(\bn, t), \bn) + \bI(\bff(\bs(\bn, t), \bn), t) \label{eq: nonlinear-intervened-solution}
\end{align}
Substitute $\bx = \bff(\bs(\bn, t) , \bn)$ into condition 1 in \Cref{def:implementation}, $\bp ( \bx + \bI(\bx, t) )= \bp(\bx) + t\bari$, and apply \Cref{eq: nonlinear-intervened-solution}:
\begin{align}
    \bp (\bs(\bn, t)) - \bp (\bff(\bs(\bn, t),\bn )) &= t\bari \label{eq:nonlinear-condition-1}
\end{align}
Differentiating \Cref{eq:nonlinear-condition-1} w.r.t. $t$ gives the first condition on $\partial_t \bs$:
\begin{align}
    \nabla\bp(\bs) - \nabla\bp\bigl(\bff(\bs,\bn)\bigr)
\nabla_{\bx}\bff(\bs,\bn) \partial_t \bs&= \bari
\end{align}

By the noise-preservation condition in \Cref{def:implementation}, we have the second condition on $\partial_t \bs$:
\begin{align}
    \nabla_{\bx} \bq(\bs, \bn) \partial_t \bs &= \mathbf 0
\end{align}
If solutions $\bs$ exist, then together with an initial condition $\bs(\bn, 0)$, $\bs$ may be solved for all of $[0, T]$ by integrating along $\partial_t \bs$. Then $\bI(\bff(\bs(\bn, T), \bn), T)$ is given by $\bs(\bn, T) - \bff(\bs(\bn, T), \bn) $.

and is well-defined under the following condition:
\begin{assumption}[Consistency of intervention values]
\label{assp:intervention-consistency}
For all $\bn,\bn'\in\mathcal{N}$,
\[
\bff(\bs(\bn, T), \bn)=\bff(\bs(\bn', T), \bn')
\quad\Longrightarrow\quad
\bs(\bn, T)
=
\bs(\bn', T)
\]
\end{assumption}

We seek an intervention field $\bi:\X\to\X$ such that at each micro-state, the direction of intervention lies in the null space of the Jacobian of $\bq$ (ensuring noise preservation) while achieving the macro-target.

The algorithm is summarised in \Cref{alg:solution-map-integration}, \Cref{app:SAI-algorithm}.
Technical conditions for the existence of implementations are in \Cref{app:nonlinear-existence-approximation}. Numerical integration approximate $\bs(\bn, T)$ to the accuracy of the corresponding solvers.

\section{Case study: machine learning model auditing and editing}
\label{sec:experiments}

In mechanistic interpretability, the micro-level causal model is
the neural network. There is growing interest in interpreting
aggregations of neurons as human-understandable variables.
For a network predicting an individual's
profession from their biography, safety auditors seek to
\emph{understand} whether it encodes a sensitive attribute
(e.g. gender) in its latent space and uses it to predict profession.
Practitioners may also \emph{control} the model by editing
its latent representation, for example to increase the model's truthfulness.
We distinguish these two goals of
mechanistic interpretability:
\begin{itemize}
    \item \textbf{Auditing} studies how model outputs depend on
    high-level concepts under input-level interventions actionable by the auditor. For example, a gender audit may examine
    how predictions change when gender cues change while other signals in the natural-language input are retained
    \citep{dearteaga2019bias,vig2020causal}.
    \item \textbf{Editing} seeks interventions that achieve a
    desired high-level effect, such as increasing truthfulness or
    changing factual associations \citep{li2023inference,zou2023representation,meng2022locating}.
    Unlike auditing, these interventions need not correspond to naturally realisable input changes: they may act directly on
    latent activations.
\end{itemize}

Our framework distinguishes auditing from editing through the
macro noise $\bq$.
For auditing, the practitioner specifies interventions on network inputs, implicitly defining a distribution
over feasible micro-level implementations.
The objective is to characterise the average response under this distribution.
Least-squares-fitting-with-intercept learns a macro model such that residuals $\bq_{\mathrm{audit}}$
have zero conditional mean under the
fitting distribution (\Cref{sec:bias-in-bios}).
For editing, implementations are selected to achieve a desired effect, without necessarily respecting the auditing distribution. They may modify internal activations directly rather than network inputs. 
Hence these implementations are invalid under the auditing model as they change $\bq_{\mathrm{audit}}$, but  may be valid under a different macro model, 
with a different mechanism and macro noise $\bq_{\mathrm{edit}}$. 
This perspective thus resolves paradoxes observed in prior work:
causal-tracing scores need not predict factual-editing success
\citep{hase2023does}, and subspace patches can activate ordinarily
dormant pathways \citep{makelov2024subspace}, we elaborate with the latter in \Cref{sec:ioi-experiment}.



\subsection{Auditing Gender Effects in Profession Classification}
\label{sec:bias-in-bios}

\paragraph{Background.}

\citet{dearteaga2019bias, vig2020causal} studied gender bias in a profession classification task where the model is asked to predict the individial's profession from their biography.
Here we take a mechanistic interpretability perspective: instead of asking how the model respond to the gender feature in the input data, we audit its response to changes in a \emph{model-internal gender
representation}.
\paragraph{Setup.}
We use a frozen RoBERTa classifier
(\texttt{Fannyjrd/roberta-bios}) that predicts 28 professions from biography
text. Gender is conveyed through textual cues, not supplied as a separate
classifier input.
Our macro cause variable is gender $G(h)$, where the aggregation map $G$ is learned by predicting the gender label from a layer's activation.\footnote{The aggregation map is often called a probe in mechinterp language.} To measure the relevance of $G(h)$ with the actual gender label of the biography, we will report the held-out accuracy of $G$. For simpliciy of exposition, we choose a macro effect variable to be the same as the micro effect variable $\mathbf Y$, which is a vector of 28 centered profession logits.

\paragraph{Experiment 1: auditing the effect of gender.} We use the dataset from \citep{dearteaga2019bias}, which contains biographies with their profession and gender labels. As we don't have access to a safety-audit practitioner, we simulate using an LLM to flip gender cues for 5000 biographies while maintaining other aspects of the text, such as length, tone, and contextual information, and we ask the LLM to `use good judgement' where needed. This protocol defines the implementation distribution used by the audit.
We split the data evenly for probe training, probe selection, macro model fitting, and held-out evaluation. For each original and edited biography in the macro-fitting split, we measure model's gender representation $G_i$ and centered-logit vector $\mathbf{Y}_i$,
and seek a macro model $\mathbf{Y}=\hat{\mathbf b}+{\bar A}G+\bq$ by fitting
\[
(\hat{\mathbf b},{\bar A})
=
\arg\min_{\mathbf b,\bar A}
\sum_i \left\|\mathbf{Y}_i-\mathbf b-\bar A G_i\right\|_2^2,
\qquad
\bq(h)=\mathbf{Y}(h)-\hat{\mathbf b}-{\bar A}G(h).
\]

\paragraph{Experiment 2: probability-level effects of gender interventions.}
To demonstrate the implication of the audit model, we intervene on gender according to the audit model found in experiment 1. We look for implementations $ h \mapsto h + i$ of gender interventions $\bar i(g) = 1 - g$. Since the macro effect variable is a vector of logits, we transform it to probabilities with $\operatorname{softmax}(\mathbf{Y}(i(h)))$ and examine the resulting female-minus-male probability changes.

\paragraph{Experiment 1 results.}
We audit the model's output dependence on its representation
of gender. A linear probe applied to the masked-mean embedding output
(layer~0, the input-embedding representation layer before the transformer blocks) achieves 99.27\%. Table~\ref{tab:bios-macro-coefficients} reports the ten largest-magnitude
coefficients among the 28 professions.
Positive coefficients indicate an increase in the centered profession
logit when decoded gender changes from male to female.
Nurse has the largest positive coefficient, with an
increase of $0.622$ centered-logit units under the specified audit
distribution. The auditor could conclude from this that the model has a mild dependence on its gender representation.

\begin{table*}[t]
\caption{Top ten fitted macro coefficients,
ordered by magnitude. Each coefficient measures the
female-minus-male change in a centered profession logit.}
\label{tab:bios-macro-coefficients}
\centering
\small
\setlength{\tabcolsep}{4pt}
\resizebox{\textwidth}{!}{%
\begin{tabular}{l*{10}{c}}
\toprule
Profession
& Nurse
& Surgeon
& Dietitian
& \shortstack{Yoga\\teacher}
& Paralegal
& Architect
& \shortstack{Interior\\designer}
& Chiropractor
& \shortstack{Software\\engineer}
& Psychologist \\
\midrule
Coefficient
& $+0.62205$
& $-0.31926$
& $+0.28134$
& $+0.26831$
& $+0.24636$
& $-0.21439$
& $+0.19793$
& $-0.18827$
& $-0.18311$
& $+0.17708$ \\
\bottomrule
\end{tabular}%
}
\end{table*}

We compare the network's responses to input-level gender edits
with the predictions of the macro model.
For each original--edited pair, we compare the rate of changes in the highest-scoring profession under input-level gender edits with what would be predicted by the macro model. For reference, we report the residual change
$\Delta\mathbf{q}$.

Results are in Table~\ref{tab:bios-text-vs-macro}. Input-level edits change
$\mathbf{q}$ by a small amount, and as a result, profession flip rates are higher for input-level edits than predicted by the macro model fixing $\bq$. This is because
prediction-flip rates are sensitive to noises in the implementations: even if the fitted coefficient were zero, 
individual residual
changes could increase a profession's relative score for some
biographies and decrease it for others 
due to textual edits to flip gender may vary in different natural language contexts.
The fitted macro coefficient characterises the average effect under the specified audit distribution, while
$\Delta\mathbf{q}$ captures implementation-dependent variation.

\begin{table}[H]
\centering
\small
\begin{tabular}{lrrr}
\toprule
Split
& \shortstack{Mean absolute\\$\Delta\mathbf{q}$ per logit}
& \shortstack{Text profession\\flips}
& \shortstack{Fixed-$\mathbf{q}$\\macro-predicted flips} \\
\midrule
Macro fit   & 0.15801 & 60/1213 & 13/1213 \\
Calibration & 0.16599 & 67/1223 & 13/1223 \\
Held-out    & 0.16016 & 65/1220 & 13/1220 \\
\bottomrule
\end{tabular}
\caption{Actual textual-edit effects versus fixed-residual macro
predictions on retained pairs. Mean absolute residual change is
averaged over pairs and all 28 logit coordinates.
A profession flip is a change in the highest-scoring class.}
\label{tab:bios-text-vs-macro}
\end{table}

\paragraph{Experiment 2 results.}
Sixteen biographies are selected:
eight at random by a deterministic hash-based sample, four with small
top-two logit margins, and four whose macro-predicted
interventions change the highest-scoring profession. We construct activation edits using Structural Alignment Integration
(\Cref{alg:solution-map-integration}).
Attainment of gender intervention and preservation of $\bq$ associated with the
binary gender macro model are verified at the final implementation. All 16 examples flip gender while preserving
$\mathbf{q}$, with maximum componentwise residual error
$1.68\times10^{-4}$. The maximum absolute discrepancy between actual and
macro-predicted class probabilities is $2.75\times10^{-5}$, demonstrating causal consistency for
these constructed interventions.
Table~\ref{tab:bios-constructed-edits} reports network
predictions before and after implementation, demonstrating
probability changes predicted by the auditting model through valid activation edits. We observe from the results that if a model is very certain on the profession entailed in a biography, the after-edit profession prediction is less likely to change; the sensitive cases are when the model is indecisive about the profession, and gender bias may result in actual profession prediction changes there (e.g. changing from physician to nurse as gender changes from male to female). We include the original and editted biographies of all 16 examples in \Cref{app:bios-constructed-examples}.

\begin{table*}[t]
\caption{Network predictions with
$\mathbf{q}$-preserving activation edits.
$G=0$ is male and $G=1$ female.
Rows with change in the predicted profession are shaded.
The probability columns describe the class with the largest
absolute probability change, which need not be the highest-scoring
class. $\Delta$ denotes after-minus-before change in percentage
points.}
\label{tab:bios-constructed-edits}
\centering
\small
\setlength{\tabcolsep}{5pt}
\resizebox{0.55\textwidth}{!}{%
\begin{tabular}{cllrr}
\toprule
$G$
& \shortstack{Predicted profession\\before $\to$ after}
& \shortstack{Largest-change\\class}
& \shortstack{Class probability\\before $\to$ after}
& $\Delta$ (pp) \\
\midrule
$1\to0$
& journalist $\to$ journalist & journalist
& $91.028\%\to90.414\%$ & $-0.614$ \\
$0\to1$
& professor $\to$ professor & professor
& $89.849\%\to91.564\%$ & $+1.715$ \\
$0\to1$
& professor $\to$ professor & professor
& $78.438\%\to75.370\%$ & $-3.068$ \\
$1\to0$
& journalist $\to$ journalist & journalist
& $95.841\%\to95.270\%$ & $-0.570$ \\
\rowcolor{professionflip}
$1\to0$
& nurse $\to$ teacher & nurse
& $25.157\%\to15.495\%$ & $-9.663$ \\
$0\to1$
& physician $\to$ physician & physician
& $49.606\%\to50.581\%$ & $+0.975$ \\
$0\to1$
& professor $\to$ professor & architect
& $0.058\%\to0.047\%$ & $-0.011$ \\
$0\to1$
& teacher $\to$ teacher & teacher
& $98.810\%\to98.877\%$ & $+0.067$ \\
$0\to1$
& physician $\to$ physician & surgeon
& $34.752\%\to29.194\%$ & $-5.558$ \\
$0\to1$
& paralegal $\to$ paralegal & paralegal
& $33.678\%\to39.158\%$ & $+5.479$ \\
\rowcolor{professionflip}
$0\to1$
& architect $\to$ professor & architect
& $49.637\%\to44.244\%$ & $-5.393$ \\
$0\to1$
& poet $\to$ poet & composer
& $36.702\%\to33.169\%$ & $-3.532$ \\
\rowcolor{professionflip}
$0\to1$
& physician $\to$ nurse & nurse
& $21.821\%\to35.026\%$ & $+13.204$ \\
\rowcolor{professionflip}
$1\to0$
& nurse $\to$ professor & nurse
& $36.739\%\to23.053\%$ & $-13.685$ \\
\rowcolor{professionflip}
$1\to0$
& nurse $\to$ pastor & nurse
& $32.864\%\to20.083\%$ & $-12.780$ \\
\rowcolor{professionflip}
$0\to1$
& rapper $\to$ model & model
& $23.135\%\to26.138\%$ & $+3.003$ \\
\bottomrule
\end{tabular}%
}
\end{table*}

\subsection{Resolving interpretability illusion: Indirect Object Identification}
\label{sec:ioi-experiment}

\paragraph{Background.}
\citet{wang2023interpretability} study indirect object identification
(IOI) in GPT-2 Small: completing ``When Mary and John went to the
store, John gave a bottle of milk to'' with ``Mary'', the indirect
object rather than the subject.
They identify ``S-Inhibition heads'' in layers 7 and 8 that write
subject-position information into the residual stream for 
``Name Mover heads'' in later layers.
\citet{makelov2024subspace} find a direction
$\mathbf{v}_{\mathrm{MLP}}$ in MLP 8's hidden activations along which
patching\footnote{Patching changes the activation's component along
$\mathbf{v}_{\mathrm{MLP}}$.} substantially changes predictions,
despite its limited IOI role on the input distribution.
This happens since $\mathbf{v}_{\mathrm{MLP}}$ combines a component
in the output projection's kernel with an output-sensitive component:
patching exploits the former's variation to activate a dormant
pathway through the latter.
The ``illusion'' is not causal ineffectiveness, but mistaking successful
editing for the MLP's feature mediation under input distribution.

\paragraph{Experiment: auditing versus editing.}
Our framework distinguishes these interpretations through the
implementation distribution and corresponding $\bq$.
Let $\bx$ denote the selected MLP hidden activation and
$\barx=\mathbf{v}_{\mathrm{MLP}}^\top\bx$.

To isolate the MLP pathway, we fix a base prompt $c$ and replace only
its final-token MLP 8 hidden activation with $\bx_i$, extracted from
each dataset prompt. The residual stream and other activation sites
at the intervention layer retain their base values; downstream
computation proceeds normally.
For each base prompt, we record the logits $\mathbf{Y}_i = (\mathrm{logit}^{(i)}_{\mathbf{subject}}, \mathrm{logit}^{(i)}_{\mathbf{object}})$ as macro effect variables and fit
\[
(\hat{\mathbf b}_c,\hat{\mathbf C}_c)
=
\arg\min_{\mathbf b,\mathbf C}
\sum_i
\left\|
\mathbf Y_i-\mathbf b-\mathbf C\,\boldsymbol{\phi}(\barx_i)
\right\|_2^2,
\qquad
\barx_i=\mathbf v_{\mathrm{MLP}}^\top\bx_i.
\]
Here $\boldsymbol{\phi}:\mathbb R\to\mathbb R^9$ contains nine
nonredundant cubic B-spline basis functions, with eight breakpoints
placed at equally spaced quantiles of the fitting coordinates,
including their minimum and maximum.
Write the fitted macro model as
\[
\mathbf{Y}_c(\barx, \bq_c)
=
\hat{\mathbf b}_c
+
\hat{\mathbf C}_c\,\boldsymbol{\phi}(\barx)
+ \bq_c
,
\qquad
\bq_c(\bx)
=
Y_c(\bx)
-
\hat{\mathbf b}_c
-
\hat{\mathbf C}_c\,
\boldsymbol{\phi}\!\left(
\mathbf v_{\mathrm{MLP}}^\top\bx
\right)
\]
which tells us the average effect of changing the macro variable $\barx$ when the change arises from natural language changes in the prompt distribution. On the other hand, we may also change the macro variable $\barx$ by changing $\bx$ along $\mathbf{v}_{\mathrm{MLP}}$. Crucially, notice that both implementations may result in the same change on the macro variable $\barx$, but may have different effects on the output logits, since the edit along  $\mathbf{v}_{\mathrm{MLP}}$ would not preserve
$\bq$.
Successful editing and a negligible audited contribution through
the same site therefore need not contradict one another.



\begin{figure*}[t]
    \centering
    \caption{
    \textbf{(a)} Subject-minus-indirect-object logit difference
    against $\barx$. Blue curves are fixed-$\bq$ macro predictions;
    blue dots are SAI implementations, one per context,
    confirming that the resulting effect is identical to macro model prediction.
    Orange curves show edits along $\mathbf{v}_{\mathrm{MLP}}$.
    \textbf{(b)} Signed changes in the auditing $\bq$ when both methods
    implement $\barx\mapsto\barx+1$.
    Orange points use
    $\bx'=\bx+\mathbf v_{\mathrm{MLP}}/\|\mathbf v_{\mathrm{MLP}}\|_2^2$.
    Blue points attain the $\barx$ target while preserving $\bq_c$.
    Orange points flip between pos. and neg. along the $y$-axis,
    matching the flipping slopes in (a); this is because
    $\mathbf{v}_{\mathrm{MLP}}$ encourages the name in the second position
    of the prompt, which is either the subject or the object
    depending on the base prompt.
    }
    \label{fig:ioi-audit-editing}
    \begin{subfigure}[t]{0.49\textwidth}
        \centering
        \includegraphics[width=\linewidth]{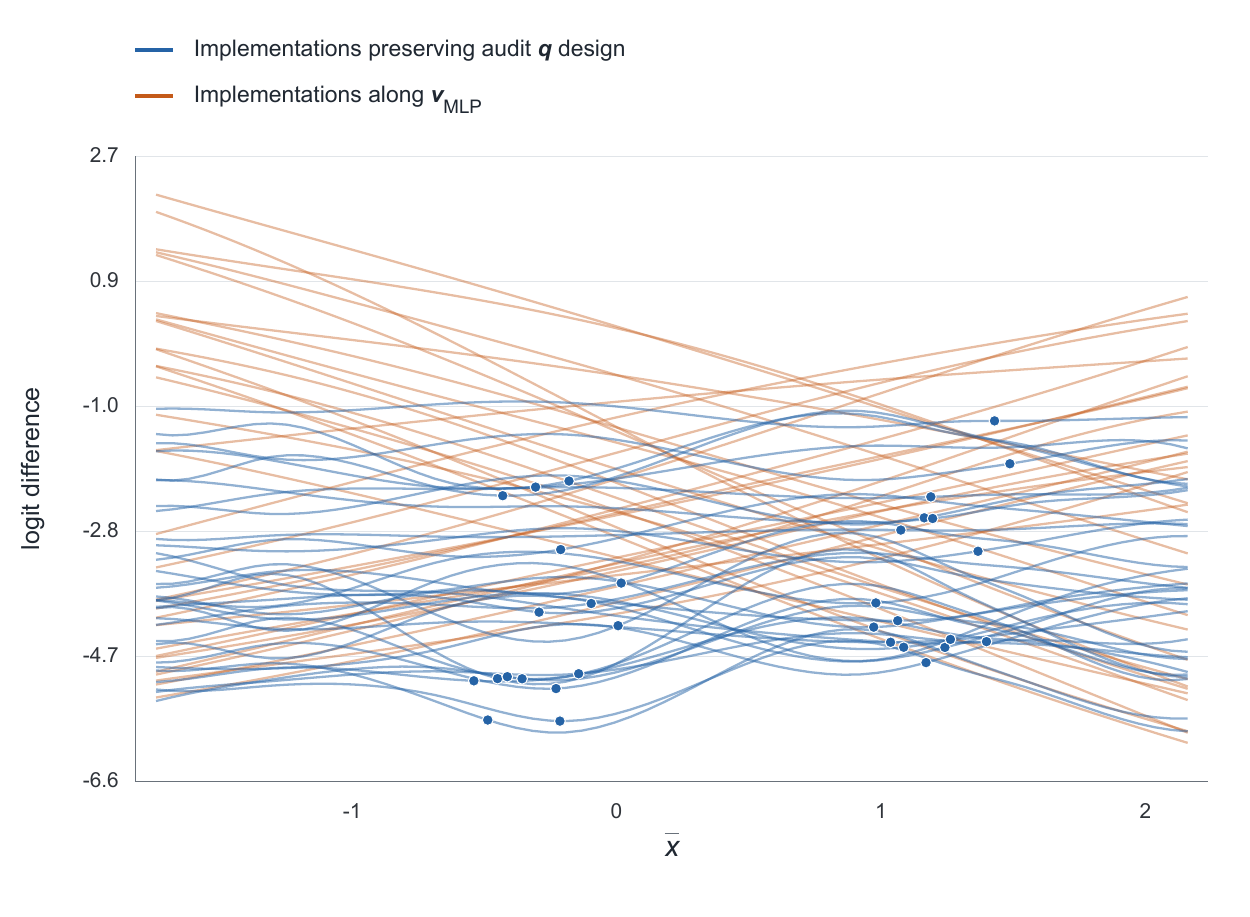}
        \caption{Fixed-$\bq$ predictions and directional edits.}
        \label{fig:ioi-audit-q-curves}
    \end{subfigure}
    \hfill
    \begin{subfigure}[t]{0.49\textwidth}
        \centering
        \includegraphics[width=\linewidth]{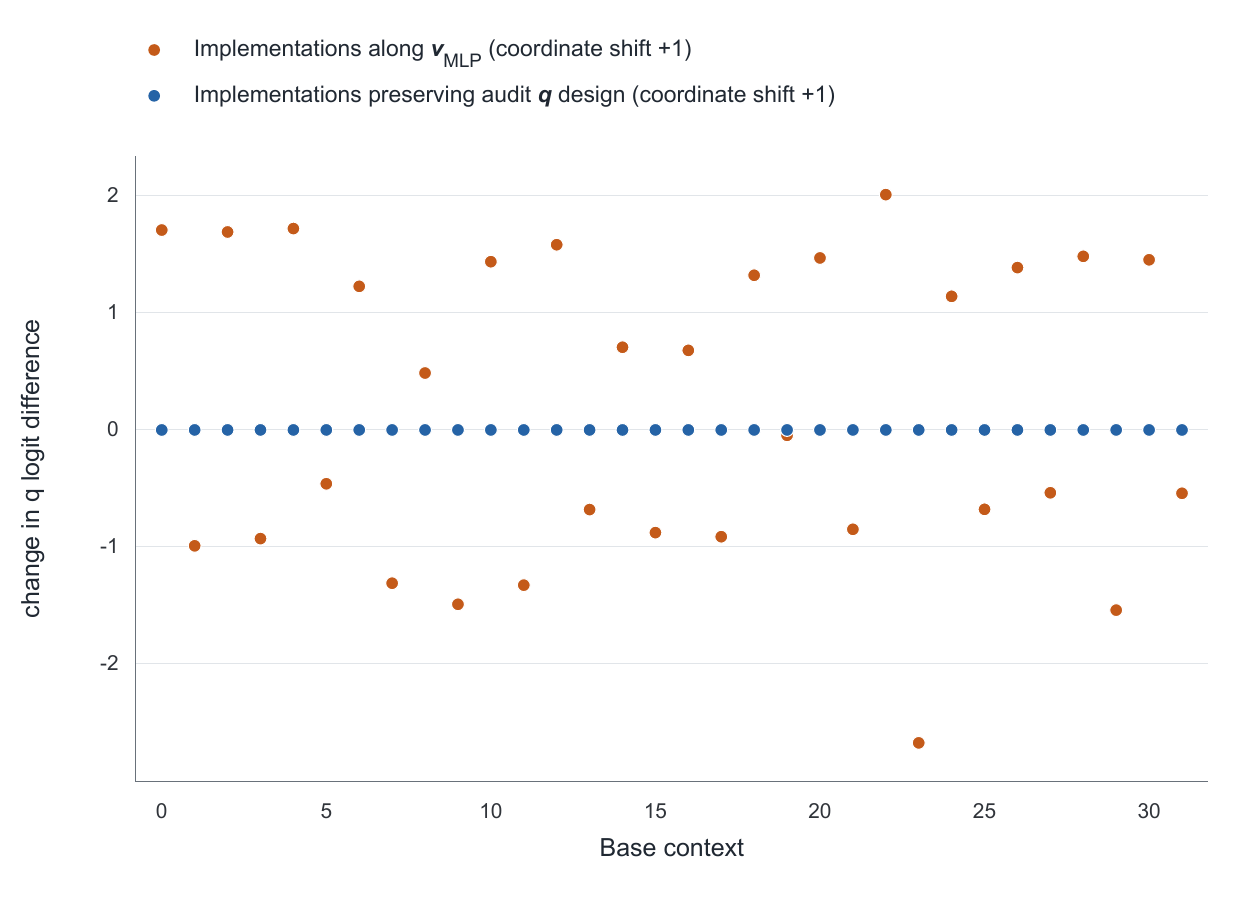}
        \caption{Signed changes in the auditing complement.}
        \label{fig:ioi-audit-q-changes}
    \end{subfigure}
\end{figure*}

\paragraph{Results.}
For each of 32 fixed base prompts, we fitted a spline auditing
model. Figure~\ref{fig:ioi-audit-q-curves} compares the predicted
logit difference between the base subject and indirect object under
interventions preserving the auditing macro noise $\bq_c$ with the
predicted logit difference responding to edits along $\mathbf v_{\mathrm{MLP}}$.
The fixed-$\bq_c$ responses show weak dependence on
$\barx$: their mean within-context range is 1.03 logits, compared with
4.01 logits for $\mathbf v_{\mathrm{MLP}}$-edits over the same coordinate
range. Constructed $\bq_c$-preserving implementations at the matched
source targets agree with the macro predictions in 32 contexts,
with maximum two-logit error $9.93\times10^{-5}$.
Figure~\ref{fig:ioi-audit-q-changes} compares changes in $\bq_c$
under edits along $\mathbf v_{\mathrm{MLP}}$ that increase $\barx$ by $+1$.
$\mathbf v_{\mathrm{MLP}}$-edits produce large changes in $\bq_c$,
which account for their departure from the fixed-$\bq_c$ macro
prediction. Their stronger effect therefore does not contradict the
weaker causal response prescribed by the auditing model: auditing and editing
are different questions and do not require the same implementations.


\vspace{-1ex}
\subsection{Discussion}
\vspace{-1ex}
In \Cref{sec:bias-in-bios}, we show a framework for auditing the model's processing mechanism of a sensitive attribute. Requiring $\bq$ to have zero mean given the macro cause allows the practitioner to square with other audits concluding a different macro model - the implementations from other audits would not leave $\bq$ in the current macro model with zero conditional mean. Then we illustrated the implications of the macro model by showing the macro coefficients and changes in the predicted professions as we implement gender interventions respecting the auditing model. In \Cref{sec:ioi-experiment}, we demonstrate that our framework is robust to a paradox observed in prior works: a valid subspace for model editing does not correspond to the subspace where the model implements the relevant signal. We show that auditing a macro variable constructed using the editing direction still correctly concludes (almost) no effect on the output. Crucially, our framework unifies editting and auditing: intervening along the editing direction changes $\bq$ under the auditing model.

\vspace{-1ex}
\section{Conclusion}
\label{sec:conclusion}
\vspace{-1ex}
We propose a complementary variable $\bq$ to remove ambiguous effects in causal abstraction, yielding a well-defined macro causal model.
Our framework shifts causal abstraction from a verificational to a constructive paradigm.
In the linear case, $\bq$ admits a unique form and we establish that under mild conditions the existence of implementations for any macro shift interventions as well as the well-definedness of causal effects. In the nonlinear case, the same uniqueness holds, and we provide an ODE-based algorithm to construct valid implementations, and demonstrate it in experiments. In the case study, we audit machine learning models with our framework and resolves a well-known paradox in mechanistic interpretability; our framework can be readily extended to explain similar paradoxes such as the mismatch between causal localisation and editing identified in \citep{hase2023does}. \footnote{Code used for the case study will be released after the manuscript is published.}


\subsection*{AI Use Statement}
We used generative AI tools for constructing gender edited biographies in \Cref{sec:bias-in-bios}. AI drafted the supplementary technical conditions included in \Cref{app:nonlinear-existence-approximation}, and we reviewed the conditions and proofs. The experiment code was written by AI and we reviewed the code. We asked AI assistants to flesh out and simplify \Cref{alg:solution-map-integration} from our ODE solution idea and under our supervision. We also used AI for discovering related works in mechanistic interpretability and we studied the papers AI found ourselves. Finally, we used AI for drafting parts of the paper from our plans and for standard text editing and polishing; we have reviewed and take responsibility of the final version of the paper. We have not used AI for developing the main ideas and framework in the paper, for designing experiments, nor for proving any of the theoretical results included in the main paper.

\bibliography{iclr2027_conference}

\bibliographystyle{iclr2027_conference}

\appendix

\section{Nonlinear implementation algorithm}
\label[appendix]{app:SAI-algorithm}
\begin{algorithm}[H]
\caption{Structural Alignment Integration via the Solution Map}
\label{alg:solution-map-integration}
\begin{algorithmic}[1]

\REQUIRE Micro mechanism $\bff$, macro noise $\bq$,
aggregation map $\bp$, initial solution $\bs_0$,
macro-shift $\bari$, and strength $T$.

\ENSURE An intervention $\bI_T$ attaining the
macro mechanism shift $T\bari$ while preserving $\bq$.

\STATE \textbf{The velocity field.}
For each state--noise pair $(\bx,\bn)$, form
\[
H(\bx,\bn)
=
\begin{bmatrix}
\nabla\bp(\bx)
-
\nabla\bp\bigl(\bff(\bx,\bn)\bigr)
\nabla_{\bx}\bff(\bx,\bn)
\\
\nabla_{\bx}\bq(\bx,\bn)
\end{bmatrix},
\qquad
\mathbf{c}
=
\begin{bmatrix}
\bari\\
\mathbf{0}
\end{bmatrix}.
\]
Select the minimum-norm feasible velocity:
\[
V(\bx,\bn)
=
\operatorname*{arg\,min}_{\mathbf{v}}
\left\{
\frac{1}{2}\|\mathbf{v}\|^2
:
H(\bx,\bn)\mathbf{v}=\mathbf{c}
\right\}.
\]

\STATE \textbf{Estimate the final solution map.}
For a micro noise $\bn$, numerically solve the ODE $\partial_t \bs(\bn, t)
=
V(\bx(t),\bn),$ with initial value $\bs_0(\bn)$ for $t\in[0,T]$.

\STATE \textbf{Return the micro-implementation.}
$\bI(\bff(\bs(\bn, T),\bn), T)
=
\bs(\bn, T)-\bff(\bs(\bn, T),\bn).$
\end{algorithmic}
\end{algorithm}
\section{Biographies and constructed gender interventions}
\label[appendix]{app:bios-constructed-examples}
We report the original biographies for all 16 selected examples. Each
``after'' prediction is obtained by applying the saved embedding intervention
to the original biography, not by substituting a text-edited biography.
The decoded gender coordinate is $G=0$ for male and $G=1$ for female.
Each row reports the logit of that row's predicted profession; when the
prediction changes, the before and after scores therefore refer to different
classes. Raw logits are denoted by $\ell_j$. Centered logits are
$Y_j=\ell_j-\frac{1}{28}\sum_{k=1}^{28}\ell_k$ and are the output
coordinates used in $q=Y-b-\bar A G$. Centering leaves softmax probabilities
unchanged. A prediction change is marked explicitly below.
These examples were deliberately selected and do not estimate a population
prediction-flip rate.
\paragraph{Example 1 (\texttt{gender-27200}).}
$G: 1\to 0$. Profession prediction unchanged.
\begin{quote}
She lives in Etna, New Hampshire. Edie Thys Morgan is a two-time Olympic skier, a former senior editor at SKI Magazine, and a freelance journalist. Her articles have appeared in O, The Oprah Magazine, Outside for Women, and Sports Illustrated for Women. She lives in Hanover, New Hampshire.
\end{quote}
\noindent\begin{tabular}{llrr}
\toprule
State & Predicted profession & Raw logit & Centered logit \\
\midrule
Before & journalist & $6.60161$ & $7.02656$ \\
After & journalist & $6.51683$ & $6.94218$ \\
\bottomrule
\end{tabular}

\paragraph{Example 2 (\texttt{gender-64323}).}
$G: 0\to 1$. Profession prediction unchanged.
\begin{quote}
He is also an Honorary Senior Lecturer at the Bartlett School of Architecture, University College London. He received a bachelor’s degree in Biology from Pennsylvania State University, a doctorate in Genetics from Yale University and was a postdoctorate fellow under Jack Szostak at Harvard University. He has published in the area of protocells, complex systems, evolution and the origin of life in various journals including JACS and Science. He is developing novel synthetic chemical systems based on the properties of living systems. These synthetic systems are termed ‘protocells’ as they are model systems of primitive living cells and chemical examples of ‘artificial’ life.
\end{quote}
\noindent\begin{tabular}{llrr}
\toprule
State & Predicted profession & Raw logit & Centered logit \\
\midrule
Before & professor & $7.82825$ & $8.26664$ \\
After & professor & $7.82278$ & $8.26971$ \\
\bottomrule
\end{tabular}

\paragraph{Example 3 (\texttt{gender-10599}).}
$G: 0\to 1$. Profession prediction unchanged.
\begin{quote}
Dr. Schielke specializes in the assessment and treatment of trauma-related and dissociative disorders, and his research focuses on increasing understanding of the treatment of these disorders and the relational components of psychotherapy. He completed a two-year postdoctoral fellowship in the assessment, differential diagnosis, and treatment of patients with complex trauma and dissociative disorders at The Trauma Disorders Program at Sheppard Pratt Health System in Baltimore, MD, and is the Co-Investigator for the Treatment of Patients with Dissociative Disorders (TOP DD) Network Study.
\end{quote}
\noindent\begin{tabular}{llrr}
\toprule
State & Predicted profession & Raw logit & Centered logit \\
\midrule
Before & professor & $7.72440$ & $8.23771$ \\
After & professor & $7.72115$ & $8.24077$ \\
\bottomrule
\end{tabular}

\paragraph{Example 4 (\texttt{gender-1075}).}
$G: 1\to 0$. Profession prediction unchanged.
\begin{quote}
She recently joined Thirteen/WNET.org “WIDE ANGLE” as a production assistant on the documentary series “Women, War \& Peace.” She previously worked as an assistant producer at Voice of America and interned at 60 Minutes and WNYC’s “The Leonard Lopate Show.” Janisch graduated from the Columbia University Graduate School of Journalism. This piece was adapted from her master’s thesis.
\end{quote}
\noindent\begin{tabular}{llrr}
\toprule
State & Predicted profession & Raw logit & Centered logit \\
\midrule
Before & journalist & $7.38996$ & $7.68091$ \\
After & journalist & $7.29712$ & $7.59654$ \\
\bottomrule
\end{tabular}

\paragraph{Example 5 (\texttt{gender-23204}).}
$G: 1\to 0$. \textbf{Profession prediction changed.}
\begin{quote}
After giving birth to her daughter she soon realized that motherhood is not as easy as it looks in the movies and spent most of her days getting covered in both baby and puppy poop, pee, and puke. Shannon started a blog as a way to document her daughter's growth and vent about the daily challenges of being a mom. Luckily, she discovered many other women share the same experiences as her and now enjoys sharing her journey with moms all over the world on her appropriately named blog PoopPeePuke.com. Shannon is originally from Kansas but now resides in New York City with her awesome husband, daughter, and dog. In her free time she likes to sleep, eat pancakes, and wait Shannon has no free time because she is a mom! That's alright though because motherhood rocks!
\end{quote}
\noindent\begin{tabular}{llrr}
\toprule
State & Predicted profession & Raw logit & Centered logit \\
\midrule
Before & nurse & $3.01095$ & $3.15919$ \\
After & teacher & $2.46487$ & $2.60749$ \\
\bottomrule
\end{tabular}

\paragraph{Example 6 (\texttt{gender-24260}).}
$G: 0\to 1$. Profession prediction unchanged.
\begin{quote}
He accepts Blue Cross Blue Shield EPO, Blue Cross Blue Shield Bronze, and Blue Cross Blue Shield HMO, as well as other insurance carriers. Dr. Golden is conversant in Spanish.
\end{quote}
\noindent\begin{tabular}{llrr}
\toprule
State & Predicted profession & Raw logit & Centered logit \\
\midrule
Before & physician & $6.45077$ & $6.57199$ \\
After & physician & $6.30170$ & $6.42219$ \\
\bottomrule
\end{tabular}

\paragraph{Example 7 (\texttt{gender-51121}).}
$G: 0\to 1$. Profession prediction unchanged.
\begin{quote}
He received his PhD from the Hong Kong University of Science and Technology. His research interests include cement chemistry, properties, structures, ecological building materials, new phosphate bonded cements, water-resistant magnesia chloride cement, and durability of concrete.
\end{quote}
\noindent\begin{tabular}{llrr}
\toprule
State & Predicted profession & Raw logit & Centered logit \\
\midrule
Before & professor & $9.28036$ & $9.73203$ \\
After & professor & $9.27139$ & $9.73508$ \\
\bottomrule
\end{tabular}

\paragraph{Example 8 (\texttt{gender-3058}).}
$G: 0\to 1$. Profession prediction unchanged.
\begin{quote}
He is known for his ability to get a reluctant learner to buy into education. His lessons have stayed with students long after graduation, and he’s often cited as students’ favorite LHS teacher. He, too, will retire this year. “Greg is a very loyal, dedicated and hardworking teacher who is committed to science education,” Glaze said. “He connects well with high school students and has motivated many students to learn more science that they would have without his inspiration. His professionalism will be missed!”
\end{quote}
\noindent\begin{tabular}{llrr}
\toprule
State & Predicted profession & Raw logit & Centered logit \\
\midrule
Before & teacher & $8.03097$ & $8.26698$ \\
After & teacher & $8.11296$ & $8.35774$ \\
\bottomrule
\end{tabular}

\paragraph{Example 9 (\texttt{gender-94131}).}
$G: 0\to 1$. Profession prediction unchanged.
\begin{quote}
He is also the super-hero known as Stars \& Garters. Stars \& Garters first encountered the Venture family when he and some other members of the CAL arrived at VenTech Tower to recruit Dr. Venture into the CAL's protection scheme. Tosh was unaware that the Ventures were already under the protection of the OSI, and he left, embarrassed.
\end{quote}
\noindent\begin{tabular}{llrr}
\toprule
State & Predicted profession & Raw logit & Centered logit \\
\midrule
Before & physician & $3.87696$ & $4.22555$ \\
After & physician & $3.72806$ & $4.07575$ \\
\bottomrule
\end{tabular}

\paragraph{Example 10 (\texttt{gender-32675}).}
$G: 0\to 1$. Profession prediction unchanged.
\begin{quote}
He has worked with prominent lawyers in raising public awareness to the dangers of asbestos. He holds a Bachelor’s degree in Finance. In his free time, he enjoys attending concerts and spending time with friends.
\end{quote}
\noindent\begin{tabular}{llrr}
\toprule
State & Predicted profession & Raw logit & Centered logit \\
\midrule
Before & paralegal & $4.07271$ & $4.14003$ \\
After & paralegal & $4.31922$ & $4.38627$ \\
\bottomrule
\end{tabular}

\paragraph{Example 11 (\texttt{gender-39465}).}
$G: 0\to 1$. \textbf{Profession prediction changed.}
\begin{quote}
He earned his Bachelor of Science in Architecture and Master of Architecture degrees from The Ohio State University and is currently a PhD student in Architecture with concentration in Design Computation at the Georgia Institute of Technology. Jeff worked as a designer in the office of Mack Scogin Merrill Elam Architects from 2002 to 2009, working in various phases of design from concept through construction and in roles from model builder to senior project manager. This semester, he is teaching second year architectural design studio at Kennesaw State University and has previously taught architecture courses at Auburn University, Georgia Institute of Technology and Southern Polytechnic State University.
\end{quote}
\noindent\begin{tabular}{llrr}
\toprule
State & Predicted profession & Raw logit & Centered logit \\
\midrule
Before & architect & $6.71517$ & $7.05107$ \\
After & professor & $6.71094$ & $7.04219$ \\
\bottomrule
\end{tabular}

\paragraph{Example 12 (\texttt{gender-11100}).}
$G: 0\to 1$. Profession prediction unchanged.
\begin{quote}
He makes his Young Masters debut in New York City at in the ‘Travellers in Time/ Focus on New Work Exhbibition’ at Site 109, LES. Adam Mysock was born in Cincinnati, Ohio in 1983 – the son of … Continue reading \ensuremath{\rightarrow}
\end{quote}
\noindent\begin{tabular}{llrr}
\toprule
State & Predicted profession & Raw logit & Centered logit \\
\midrule
Before & poet & $4.51052$ & $4.65102$ \\
After & poet & $4.54399$ & $4.68889$ \\
\bottomrule
\end{tabular}

\paragraph{Example 13 (\texttt{gender-13492}).}
$G: 0\to 1$. \textbf{Profession prediction changed.}
\begin{quote}
He is the co-founder of the Abstinence Clearinghouse, and served on the South Dakota Task Force to Study Abortion. He is a member of the Vote Yes For Life Campaign.
\end{quote}
\noindent\begin{tabular}{llrr}
\toprule
State & Predicted profession & Raw logit & Centered logit \\
\midrule
Before & physician & $3.72353$ & $4.10760$ \\
After & nurse & $4.22029$ & $4.60673$ \\
\bottomrule
\end{tabular}

\paragraph{Example 14 (\texttt{gender-84032}).}
$G: 1\to 0$. \textbf{Profession prediction changed.}
\begin{quote}
She is also the co-founder of Suriname Indigenous Health Fund. Sarah led a team of Indigenous and church leaders to draft the World Council of Churches (WCC) Statement on the Doctrine of Discovery and its enduring impact on Indigenous Peoples, which was adopted in 2012, and organized an international coalition to deepen institutional commitment to an Indigenous-led Program at the WCC. Sarah is working within an international ecumenical movement to dismantle the Doctrine of Discovery and to build a solidarity movement promoting dialogue with resource extractive industries among people of faith. She is a member of Seattle Mennonite Church.
\end{quote}
\noindent\begin{tabular}{llrr}
\toprule
State & Predicted profession & Raw logit & Centered logit \\
\midrule
Before & nurse & $4.38941$ & $4.78116$ \\
After & professor & $4.21909$ & $4.60897$ \\
\bottomrule
\end{tabular}

\paragraph{Example 15 (\texttt{gender-18038}).}
$G: 1\to 0$. \textbf{Profession prediction changed.}
\begin{quote}
Melissa’s passions include worship, spiritual formation, outdoor adventures, health and fitness, wilderness medicine, and serving as mom to her two vivacious children.
\end{quote}
\noindent\begin{tabular}{llrr}
\toprule
State & Predicted profession & Raw logit & Centered logit \\
\midrule
Before & nurse & $3.66052$ & $3.79442$ \\
After & pastor & $3.63947$ & $3.76770$ \\
\bottomrule
\end{tabular}

\paragraph{Example 16 (\texttt{gender-11542}).}
$G: 0\to 1$. \textbf{Profession prediction changed.}
\begin{quote}
In response to a hateful comment made about him, Ross recorded an empowering video encouraging youth to unapologetically be true to who they are, and to respond to messages of hate with messages of love. It Gets Better Project
\end{quote}
\noindent\begin{tabular}{llrr}
\toprule
State & Predicted profession & Raw logit & Centered logit \\
\midrule
Before & rapper & $3.23067$ & $3.29494$ \\
After & model & $3.17316$ & $3.24885$ \\
\bottomrule
\end{tabular}

\section{Theoretical Results}
\label[appendix]{app:theoretical-results}
\subsection{Linear closed form of macro noise}
\label[appendix]{app:proof-linear-q}

\begin{proof}[Proof of \Cref{prop:q-form}]
The kernel condition implies that the rows of $\begin{pmatrix}\bq_x & \bq_n\end{pmatrix}$ lie in the row space of $\begin{pmatrix} \bP & 0 \end{pmatrix}$ and $\bP\begin{pmatrix}\bA & I\end{pmatrix}$.
Hence $\begin{pmatrix}\bq_x & \bq_n\end{pmatrix} = B\begin{pmatrix}\bP & 0\end{pmatrix} + C\bP\begin{pmatrix}\bA & I\end{pmatrix}$ for matrices $B, C$.
Comparing the noise block gives $\bq_n = C\bP$; since $\bP$ is full-rank, $C$ must be invertible.
Requiring the macro model to have the form $\begin{pmatrix}\barA & I\end{pmatrix}$ fixes $C = I$ and $B = -\barA$, yielding~\eqref{eq:q-linear}.
\end{proof}

\subsection{Well-definedness of linear macro-interventions}
\label[appendix]{app:proof-linear-welldef}

\begin{proof}[Proof of \Cref{thm:linear-welldef}]
Let $\bi$ be any implementation of $\bari$ with respect to $(\bp, \bq)$.
By definition:
\begin{align*}
\bp\circ\bs^{(\bi)}(\bn) 
&= \bp\circ\bi(\bA\bs^{(\bi)}(\bn) + \bn) 
= \bari\circ\bp(\bA\bs^{(\bi)}(\bn) + \bn) \\
&= \bari\left(\bP\bA\bs^{(\bi)}(\bn) + \bP\bn\right) \\
&= \bari\left(\bq(\bs^{(\bi)}(\bn), \bn) + \barA\bP\bs^{(\bi)}(\bn)\right) \\
&= \bari\left(\barA\bp\circ\bs^{(\bi)}(\bn) + \bq(\bs^{(\bi)}(\bn), \bn)\right).
\end{align*}
The fourth equality uses the construction of $\bq$: $\bq(\bx, \bn) = (\bP\bA - \barA\bP)\bx + \bP\bn = \bP\bA\bx + \bP\bn - \barA\bP\bx$.

This shows that $\bp\circ\bs^{(\bi)}(\bn)$ is a solution to the intervened macro SCM $\barx = \bari(\barA\barx + \barn)$ with noise $\barn = \bq(\bs^{(\bi)}(\bn), \bn) = \bq(\bs(\bn), \bn)$ (the last equality by criterion~2 of implementation).
By \Cref{assp:unique-solutions}, this solution is unique, so $\bp\circ\bs^{(\bi)}(\bn)$ is the same for all implementations $\bi$.
\end{proof}

\subsection{Existence of macro shift interventions}
\label[appendix]{app:proof-linear-existence}

\begin{proof}[Proof of \Cref{thm:linear-existence}]
By \Cref{def:implementation}, a micro-level shift $\bi(\bx) = \bx + \bm\delta$ implements $\bari$ if and only if
:
\begin{align}
\bP\bm\delta &= \bar{\bm c}, \label[appendix]{eq:exist-1} \\
(\bP\bA - \barA\bP)(I - \bA)^{-1}\bm\delta &= \bm 0. \label[appendix]{eq:exist-2}
\end{align}
Together these impose $2\bar k$ linear constraints on $\bm\delta\in\R^k$.
Under \Cref{assp:dim}, $k \geq 2\bar k$, and \Cref{assp:lindep} ensures the $2\bar k$ constraints are independent.
Hence the system has a solution space of dimension $k - 2\bar k > 0$.
\end{proof}

\subsection{Well-definedness of nonlinear macro-interventions}
\label[appendix]{app:proof-nonlinear-welldef}

\begin{proof}[Proof of \Cref{thm:nonlinear-welldef}]
The proof follows the same structure as \Cref{thm:linear-welldef}:
\begin{align*}
\bp\circ\bs^{(\bi)}(\bn) 
&= \bari\circ\bp\circ\bff(\bs^{(\bi)}(\bn), \bn) \\
&= \bari\circ\barf(\bp(\bs^{(\bi)}(\bn)), \bq(\bs^{(\bi)}(\bn), \bn)) \\
&= \bari\circ\barf(\bp(\bs^{(\bi)}(\bn)), \bq(\bs(\bn), \bn)),
\end{align*}
where the last equality uses the noise-preservation criterion.
This shows $\bp\circ\bs^{(\bi)}(\bn)$ solves the intervened macro SCM with noise $\bq(\bs(\bn), \bn)$; uniqueness (\Cref{assp:unique-solutions}) gives invariance over implementations.
\end{proof}

\subsection{Existence and approximation of implementations in the nonlinear regime}
\label[appendix]{app:nonlinear-existence-approximation}
\begin{assumption}[Injectivity of the mechanism in noise]
\label[appendix]{assp:noise-injectivity}
For every fixed $\bx\in\mathbb{R}^{k}$, the map
$\bn\mapsto\bff(\bx,\bn)$ is injective. That is,
for all $\bn,\bn'\in\mathcal{N}$,
\[
\bff(\bx,\bn)=\bff(\bx,\bn')
\quad\Longrightarrow\quad
\bn=\bn'.
\]
\end{assumption}

\begin{theorem}[Existence of valid implementations]
\label[appendix]{thm:nonlinear-implementation-existence}
Let $\mathcal{N}\subseteq\mathbb{R}^{d}$ be open,
and let the micro-state space be $\mathbb{R}^{k}$.
Suppose that $\bff$, $\bp$, and $\bq$ are twice
continuously differentiable, and that
$\bs_0:\mathcal{N}\to\mathbb{R}^{k}$ is continuously
differentiable and satisfies
\[
\bs_0(\bn)=\bff(\bs_0(\bn),\bn).
\]
Fix a macro-shift direction $\bari$ and strength $T>0$.
Let $H$ and $\mathbf{c}$ be as defined in
\Cref{alg:solution-map-integration}.
Assume:
\begin{enumerate}
    \item \textbf{Regularity and feasibility.}
    The matrix $H(\bx,\bn)$ has constant rank, and
    \[
    \mathbf{c}\in\operatorname{range}H(\bx,\bn)
    \]
    for every
    $(\bx,\bn)\in\mathbb{R}^{k}\times\mathcal{N}$.

    \item \textbf{Controlled growth.}
    The minimum-norm velocity
    \[
    V(\bx,\bn)=H(\bx,\bn)^\dagger\mathbf{c}
    \]
    satisfies, for every compact
    $K\subseteq\mathcal{N}$,
    \[
    \|V(\bx,\bn)\|
    \leq a_K+b_K\|\bx\|
    \]
    for constants $a_K,b_K\geq0$, uniformly over
    $\bx\in\mathbb{R}^{k}$ and $\bn\in K$.
\end{enumerate}

Then the initial-value problem
\[
\partial_t\bs(\bn,t)
=
V(\bs(\bn,t),\bn),
\qquad
\bs(\bn,0)=\bs_0(\bn),
\]
has a unique solution on $[0,T]$ for each $\bn$.
It satisfies
\[
\bp(\bs(\bn,t))
-
\bp\bigl(\bff(\bs(\bn,t),\bn)\bigr)
=
t\bari,
\]
and
\[
\bq(\bs(\bn,t),\bn)
=
\bq(\bs_0(\bn),\bn).
\]

Define
\[
\bs_T(\bn)=\bs(\bn,T),
\qquad
g_T(\bn)=\bff(\bs_T(\bn),\bn),
\qquad
\mathcal{U}_T=g_T(\mathcal{N}).
\]
If \Cref{assp:intervention-consistency} holds, then
\[
\bI_T(g_T(\bn))
=
\bs_T(\bn)-g_T(\bn)
\]
defines a single-valued intervention
$\bI_T:\mathcal{U}_T\to\mathbb{R}^{k}$ satisfying
\[
\bp\bigl(\mathbf{u}+\bI_T(\mathbf{u})\bigr)
=
\bp(\mathbf{u})+T\bari
\qquad
\text{for all }\mathbf{u}\in\mathcal{U}_T,
\]
and
\[
\bs_T(\bn)
=
\bff(\bs_T(\bn),\bn)
+
\bI_T\bigl(\bff(\bs_T(\bn),\bn)\bigr).
\]
Under \Cref{assp:noise-injectivity}, $\bs_T$ is the
intervened solution map. Thus $\bI_T$ is a valid
implementation on $\mathcal{U}_T$.

If implementations are required on the entire
mechanism-output domain, additionally require that
$\mathcal{U}_T$ equals that domain, or that $\bI_T$
admits an extension satisfying the same alignment
condition there.
\end{theorem}

\begin{proof}
The constant-rank assumption implies that $H^\dagger$
is continuously differentiable. Consequently,
$V=H^\dagger\mathbf{c}$ is locally Lipschitz in $\bx$,
giving local existence and uniqueness of the ODE solution.

For fixed $\bn\in K$, the growth bound gives
\[
\|\bs(\bn,t)\|
\leq
\|\bs_0(\bn)\|
+
\int_0^t
\left(a_K+b_K\|\bs(\bn,\tau)\|\right)d\tau.
\]
Gronwall's inequality implies, when $b_K>0$,
\[
\|\bs(\bn,t)\|
\leq
e^{b_Kt}\|\bs_0(\bn)\|
+
\frac{a_K}{b_K}(e^{b_Kt}-1).
\]
When $b_K=0$, the bound is
$\|\bs_0(\bn)\|+a_Kt$.
Thus trajectories remain bounded on finite time
intervals and extend to $[0,T]$
\citep[Theorem~2.17]{teschl2012ordinary}.

Feasibility gives $HV=\mathbf{c}$.
By the chain rule,
\[
\frac{d}{dt}
\left[
\bp(\bs(\bn,t))
-
\bp\bigl(\bff(\bs(\bn,t),\bn)\bigr)
\right]
=
\bari,
\]
and
\[
\frac{d}{dt}\bq(\bs(\bn,t),\bn)=\mathbf{0}.
\]
The initial structural equation makes the bracketed
quantity zero at $t=0$. Integration therefore gives
\[
\bp(\bs_T(\bn))
-
\bp\bigl(\bff(\bs_T(\bn),\bn)\bigr)
=
T\bari,
\]
and
\[
\bq(\bs_T(\bn),\bn)=\bq(\bs_0(\bn),\bn).
\]

Write
\[
g_T(\bn)=\bff(\bs_T(\bn),\bn),
\qquad
\mathcal{U}_T=g_T(\mathcal{N}).
\]
For $\mathbf{u}\in\mathcal{U}_T$, choose any $\bn$
such that $g_T(\bn)=\mathbf{u}$ and define
\[
\bI_T(\mathbf{u})=\bs_T(\bn)-\mathbf{u}.
\]
By \Cref{assp:intervention-consistency}, this value
is independent of the chosen representative.
The preceding identities imply
\[
\bp\bigl(\mathbf{u}+\bI_T(\mathbf{u})\bigr)
=
\bp(\mathbf{u})+T\bari,
\]
and
\[
\bs_T(\bn)
=
\bff(\bs_T(\bn),\bn)
+
\bI_T\bigl(\bff(\bs_T(\bn),\bn)\bigr).
\]
Thus $\bs_T(\bn)$ is an intervened solution.

It remains to establish uniqueness.
Fix $\bn$ and suppose $\bx'$ is another solution,
with $\mathbf{u}'=\bff(\bx',\bn)\in\mathcal{U}_T$.
Choose $\bn'\in\mathcal{N}$ such that
$g_T(\bn')=\mathbf{u}'$.
By the definition of $\bI_T$,
\[
\bx'
=
\mathbf{u}'+\bI_T(\mathbf{u}')
=
\bs_T(\bn').
\]
Consequently,
\[
\bff(\bx',\bn)
=
\mathbf{u}'
=
g_T(\bn')
=
\bff(\bs_T(\bn'),\bn')
=
\bff(\bx',\bn').
\]
By \Cref{assp:noise-injectivity}, $\bn=\bn'$,
and hence $\bx'=\bs_T(\bn)$.
Therefore $\bs_T$ is the unique solution map
on the constructed mechanism-output domain.
Together with alignment and noise preservation,
this establishes validity of $\bI_T$.
\end{proof}

\begin{theorem}[Approximation accuracy]
\label[appendix]{thm:nonlinear-implementation-accuracy}
Suppose the conditions of
\Cref{thm:nonlinear-implementation-existence} hold,
and fix a compact noise domain $K\subseteq\mathcal{N}$.

Let $\widehat{\bs}_T$ approximate the terminal solution
map with uniform error
\[
\sup_{\bn\in K}
\|\widehat{\bs}_T(\bn)-\bs_T(\bn)\|
\leq \varepsilon_s.
\]
Assume that, on the relevant domains,
\[
\|\bff(\bx,\bn)-\bff(\bx',\bn)\|
\leq L_f\|\bx-\bx'\|,
\]
\[
\|\bs_T(\bn)-\bs_T(\bn')\|
\leq L_s\|\bn-\bn'\|,
\]
and
\[
\|\bn-\bn'\|
\leq
\kappa\|g_T(\bn)-g_T(\bn')\|
\qquad
\text{for all }\bn,\bn'\in K.
\]
Define
\[
\widehat g_T(\bn)
=
\bff(\widehat{\bs}_T(\bn),\bn).
\]
For an input $\mathbf{u}\in g_T(K)$, suppose the
numerical inverse solver returns $\widehat{\bn}\in K$
satisfying
\[
\|\widehat g_T(\widehat{\bn})-\mathbf{u}\|
\leq\varepsilon_{\mathrm{inv}},
\]
and define
\[
\widehat{\bI}_T(\mathbf{u})
=
\widehat{\bs}_T(\widehat{\bn})-\mathbf{u}.
\]
Then
\[
\boxed{
\|\widehat{\bI}_T(\mathbf{u})-\bI_T(\mathbf{u})\|
\leq
(1+L_s\kappa L_f)\varepsilon_s
+
L_s\kappa\varepsilon_{\mathrm{inv}}.
}
\]
The bound is uniform over inputs for which the stated
inverse-solver condition holds.

In particular, explicit Euler integration with
$h=T/N$ and exact initialization satisfies, for
sufficiently small $h$,
\[
\max_{0\leq j\leq N}
\sup_{\bn\in K}
\|\widehat{\bs}_{jh}(\bn)-\bs_{jh}(\bn)\|
\leq C_s h,
\]
where $C_s$ is independent of $h$. Consequently,
\[
\sup_{\mathbf{u}}
\|\widehat{\bI}_T(\mathbf{u})-\bI_T(\mathbf{u})\|
=
O(h+\varepsilon_{\mathrm{inv}}).
\]
More generally, a uniform solution-map error
$O(h^\ell)$ yields intervention error
$O(h^\ell+\varepsilon_{\mathrm{inv}})$.
\end{theorem}

\begin{proof}
Write $\bn=g_T^{-1}(\mathbf{u})$.
The forward approximation satisfies
\[
\|\widehat g_T(\widehat{\bn})
      -g_T(\widehat{\bn})\|
\leq L_f\varepsilon_s.
\]
Therefore,
\[
\begin{aligned}
\|\widehat{\bn}-\bn\|
&\leq
\kappa\|g_T(\widehat{\bn})-g_T(\bn)\|\\
&\leq
\kappa\bigl(
L_f\varepsilon_s+\varepsilon_{\mathrm{inv}}
\bigr).
\end{aligned}
\]
Using the definitions of the two interventions,
\[
\begin{aligned}
\|\widehat{\bI}_T(\mathbf{u})-\bI_T(\mathbf{u})\|
&=
\|\widehat{\bs}_T(\widehat{\bn})-\bs_T(\bn)\|\\
&\leq
\varepsilon_s+L_s\|\widehat{\bn}-\bn\|\\
&\leq
(1+L_s\kappa L_f)\varepsilon_s
+
L_s\kappa\varepsilon_{\mathrm{inv}}.
\end{aligned}
\]

For the Euler estimate, continuous dependence on the
noise and compactness of $K\times[0,T]$ imply that the
exact trajectories lie in a common compact set.
The velocity field is continuously differentiable on
a neighbourhood of this set. Thus its spatial
Lipschitz constant and the local truncation-error
constant can be chosen uniformly over $K$.
The Euler error recursion and a discrete Gronwall
bound give the uniform $O(h)$ estimate.
Substitution into the intervention bound completes
the proof.
\end{proof}

\end{document}